# Lightweight framework for underground pipeline recognition and spatial localization based on multi-view 2D GPR images

Haotian Lv, Chao Li, Jiangbo Dai, Yuhui Zhang, Zepeng Fan, Yiqiu Tan, Dawei Wang*, Binglei Xie

*Abstract*—To address the issues of weak correlation between multi-view features, low recognition accuracy of small-scale targets, and insufficient robustness in complex scenarios in underground pipeline detection using three-dimensional ground penetrating radar (3D GPR), this paper proposes a three-dimensional pipeline intelligent detection framework that integrates multi-strategy improved deep learning technology. This paper explores a novel pathway to achieve accurate 3D localization through lightweight joint analysis of multi-view 2D GPR images. First, based on a B/C/D-Scan three-view joint analysis strategy, a three-dimensional pipeline three-view feature evaluation method is established by cross-validating forward simulation results obtained using time-domain finite difference (FDTD) methods with actual measurement data. Second, the DCO-YOLO framework is proposed, which integrates DySample dynamic upsampling, Convolutional gate linear unit (CGLU), and OutlookAttention cross-dimensional correlation mechanisms into the original YOLOv11 algorithm, significantly improving the small-scale pipeline edge feature extraction capability. Furthermore, a 3D-DIoU spatial feature matching algorithm is proposed, which integrates three-dimensional geometric constraints and center distance penalty terms to achieve automated association of multi-view annotations. The three-view fusion strategy resolves inherent ambiguities in single-view detection. Experiments based on 100 kilometers of real urban underground pipeline data show that the proposed method achieves accuracy, recall, and mean average precision of 96.2%, 93.3%, and 96.7%, respectively, in complex multi-pipeline scenarios, which are 2.0%, 2.1%, and 0.9% higher than the baseline model. Ablation experiments validated the synergistic optimization effect of the dynamic feature enhancement module, and Grad-CAM++ heatmap visualization demonstrated that the improved model significantly enhanced its ability to focus on pipeline geometric features. This study integrates deep learning optimization strategies with the physical characteristics of 3D GPR, offering an efficient and reliable novel technical framework for the intelligent recognition and localization of underground pipelines.

This work was supported by the National Key Research and Development Program of China [grant number 2023YFB2603500]; Project of Shenzhen Science and Technology Plan [grant number KJZD20230923115206014]; Heilongjiang Natural Science Foundation Research Team Project [grant number TD2022E001]. *(Corresponding author: Dawei Wang).*

Haotian Lv, Chao Li, Jiangbo Dai, Yuhui Zhang, Zepeng Fan, Yiqiu Tan and Dawei Wang are with the School of Transportation Science and Engineering, Harbin Institute of Technology, Harbin 150090, China. (e-mail: lvhaotian@stu.hit.edu.cn; 2021110450@stu.hit.edu.cn; 24b932009@stu.hit.edu.cn; yuhuizhang@stu.hit.edu.cn; zepeng.fan@hit.edu.cn; tanyiqiu@hit.edu.cn; dawei.wang@hit.edu.cn).

Binglei Xie is with the School of Architecture, Harbin Institute of Technology (Shenzhen), Pingshan First Road 6, Shenzhen, 518000, China. (e-mail: xiebinglei@126.com).

Color versions of one or more of the figures in this article are available online at http://ieeexplore.ieee.org

## I. INTRODUCTION

URBAN underground utility networks are critical infrastructure for cities, serving as the lifeline that sustains essential services such as water supply, drainage, gas distribution, and telecommunications. However, issues such as aging pipeline materials, inadequate planning and design, complex load-environment interactions, and poor maintenance and management have led to pipeline leaks, misalignments, and deformations, ultimately causing severe consequences such as road collapses and urban flooding [1]. Therefore, efficient and precise detection technologies are needed to rapidly identify the distribution and operational status of underground pipelines, thereby providing a basis for their health assessment and maintenance planning. This is an essential approach to extending pipeline service life, preventing road collapses, reducing economic losses, and ensuring public safety.

Traditional underground pipeline inspection techniques include manual visual inspection, quick view, closed-circuit television, sonar and panoramic laser detection. These methods require inspection equipment to be inserted into the pipeline, resulting in poor environmental adaptability, low inspection efficiency, complex operations, and high costs [2]. In contrast, ground penetrating radar (GPR) is a non-destructive inspection technology that emits high-frequency electromagnetic waves and analyzes the reflected signals to achieve non-destructive detection of pipelines. This method has the advantages of wide applicability, high resolution, high efficiency and safety. It can not only identify pipelines but also surrounding soil defects and structural defects, and comprehensively assess the health status of pipelines [3, 4]. However, its single-antenna structure results in low spatial resolution and difficulty in resolving complex pipeline routes. Three-dimensional ground penetrating radar (3D GPR), utilizing multi-channel array antennas, can simultaneously collect B/C/D-scan data, further enhancing detection accuracy and efficiency. This technology enables high-resolution and wide-coverage detection, making the 3D morphology of pipelines and surrounding geological structures visually presentable [5]. However, 3D GPR generates a large volume of data. Manual interpretation is time-consuming, subjective, and prone to errors. Therefore, there is an urgent need to integrate artificial intelligence technology to achieve automatic data analysis and underground target recognition, thereby improving data processing accuracy and efficiency.

The application of deep learning algorithms to 3D GPR data has witnessed remarkable progress, driven by significant advancements in model architectures and data processing techniques. Researchers such as Li et al. [6], Lv et al. [7], Huang et al. [8], and Shen et al. [9] have conducted in-depth analyses of GPR image characteristics, leading to innovative improvements in existing model architectures. For instance, Kim et al. introduced a multi view object detection method that integrates B-scan and C-scan data, significantly enhancing classification accuracy in urban road scenarios [10]. Du et al. demonstrated the effectiveness of BP neural networks in leveraging multi-scale gray-level co-occurrence matrix attributes for road hazard recognition[11]. Khudoyarov et al. developed a 3D convolutional neural network (CNN) that achieves an accuracy of up to 97% in classifying pipes, cavities, and manholes using voxel-based 3D GPR data [12].

Further architectural innovations include the 3D Res-Attention UNet [13], which combines residual connections and attention mechanisms to improve segmentation precision for pipe detection, the BALF detector [14] for robust local feature representation in challenging imaging conditions, and methods incorporating open-scan clustering algorithms with parabolic fitting-based judgment for enhanced target identification [15]. Additionally, Tong et al. proposed an evidential transformer model for robust 3D object detection under uncertainty [5]. Collectively, these studies mark a shift from traditional 2D analysis towards 3D deep learning frameworks, enabling automated and high-precision interpretation of subsurface features.

In terms of data processing methodologies, systematic preprocessing of GPR data is essential to address inherent noise and structural complexities. Lei et al. employed clustering algorithms and data augmentation techniques to enhance feature extraction from B-scan images [16]. Kim et al. implemented data chunking and grid transformation strategies to convert 3D volumetric data into 2D formats compatible with conventional CNNs [17]. Yamaguchi et al. effectively combined Kirchhoff migration with a 3D CNN for precise spatial localization of subsurface pipes [18]. Cheng et al. introduced a neural network framework designed to interpolate sparse B-scan sequences into dense 3D representations, thereby improving the reconstruction accuracy of pipes and cavities [19]. Pham et al. automated hyperbola detection using a Faster R-CNN model augmented with data synthesis techniques [20], while Xiong et al. leveraged YOLOv3 and U-Net architectures for the automatic fitting and localization of irregular hyperbolic signatures [21]. To improve signal quality, Li et al. proposed clutter suppression techniques based on the BM3D algorithm for effective noise reduction [22]. These diverse approaches, spanning signal denoising, feature enhancement, and dimensional adaptation, form a comprehensive pipeline for efficient GPR data utilization.

Future research directions emphasize addressing challenges such as data scarcity and improving model generalization across diverse and complex field conditions. To mitigate data insufficiency, researchers have extensively explored data augmentation techniques, including forward modeling [4, 23, 24], adversarial training [25, 26], and transfer learning [27]. Forward modeling methods, such as the time-domain finite difference (FDTD) technique, generate synthetic GPR data by simulating electromagnetic wave propagation for training deep learning models. For instance, Raja et al. proposed integrating FDTD with optimization algorithms to estimate pipe properties, enhancing the physical realism of simulated data [28]. Adversarial training leverages generative adversarial network (GAN) to enhance data diversity. Xu et al. developed a coupled-learning GAN framework that learns features from both simulated and real data through a dual-stage training strategy, improving the robustness of GPR image inversion [29]. Transfer learning adapts pre-trained models to new scenarios, reducing reliance on large-scale annotated data. Dai et al. employed a transfer learning strategy, applying a pre-trained network to 3D GPR data inversion under heterogeneous soil conditions, improving model generalization with limited data[30]. While these methods partially alleviate data scarcity, they often diverge from real-world conditions. However, the idealized nature of simulated data limits model performance in practical deployments. Cui et al. emphasized the urgent need for large-scale, annotated real-world datasets to reduce dependence on simulated data, as simulations frequently fail to capture the full complexity of field conditions [31]. Pei et al. constructed the standardized real-world dataset based on 3D GPR, validating the critical role of real data in improving defect detection accuracy [32]. Liu et al., in a comprehensive review, noted that while simulated data can be used for preliminary training, model performance in real road scenarios heavily depends on the scale and diversity of annotated real data [33]. Furthermore, Wang et al. experimentally demonstrated that models trained solely on simulated data exhibit high false-positive rates in complex underground environments, whereas the introduction of real annotated data significantly enhances detection robustness [34]. Collectively, these studies confirm that while data augmentation methods using simulations offer certain benefits, they cannot replace the diversity and noise characteristics inherent in real-world data. In conclusion, large-scale, annotated real-world datasets constitute the most fundamental foundation for deploying trained models into practical applications, enabling maximum robustness and environmental adaptability.

Current deep learning-based intelligent recognition methods for 3D GPR data can be broadly categorized into two main approaches: those based on 3D CNN and those utilizing multi-view data. The 3D CNN based approach processes the 3D volumetric data directly. For instance, Khudoyarov et al. proposed a 3D GPR data classification network based on a 3D CNN architecture for classifying underground objects such as pipes and cavities [12]. Yamaguchi et al. [35] and Zhou et al. [36] developed models that integrate 3D CNN with Kirchhoff migration for the detection and 3D localization of pipes. The multi-view data-based approach projects or decomposes the 3D data into multiple 2D views. Huang et al. proposed a multi-perspective cascade recognition method that effectively integrates B-scan and C-scan images to enhance the recognition of pavement subsurface objects [8]. Kim et al. employed 3D GPR data as 2D grid images from B, C, and D-scan views for underground object classification [10, 17].

These two approaches have advantages and disadvantages. The 3D CNN based method can preserve the spatial 3D correlations of the original data and possesses strong theoretical feature extraction capabilities. However, its drawbacks are significant: (1) High computational complexity and memory demand, as processing the entire 3D data volume hinders real-time deployment in practical applications; (2) Redundant feature extraction, as the high similarity between adjacent slices may lead the model to learn repetitive features, reducing its discriminative power; (3) Insufficient accuracy for small-scale targets, where operations like pooling can degrade fine-grained spatial details, impairing the identification of small objects [12]. In contrast, the advantages of the multi-view data-based approach include: (1) High data processing efficiency, as converting 3D data into 2D images significantly reduces data volume and computational complexity, leading to simpler and faster algorithms; (2) Strong compatibility, allowing direct use of mature 2D CNN architectures and pre-trained models. However, the core challenge of this method lies in fully exploiting and integrating the complementary features from different views. Although existing research has demonstrated its potential, it is still in a developmental stage and has not yet systematically leveraged the deep feature relationships and correlations between different views of GPR images [8].

In summary, the multi-view data-based approach shows great potential in computational efficiency and practicality, representing a crucial direction for advancing 3D GPR technology towards engineering applications. Therefore, this study proposes a novel solution: it pioneers a new pathway for achieving precise 3D detection of underground pipelines by jointly analyzing 2D multi-view data (B/C/D-scan), effectively circumventing the inherent challenges of high computational complexity and memory demands associated with direct processing of 3D data volumes. Furthermore, a new intelligent recognition network, DCO-YOLO, is constructed based on the YOLOv11 framework. By integrating modules such as DySample, Convolutional gate linear unit (CGLU), and OutlookAttention, the network significantly enhances the characterization of small-scale target edge features and cross-dimensional feature correlations. Experimental results demonstrate effective improvement in key performance metrics within complex scenarios, offering a new method for the intelligent interpretation of 3D GPR data that balances both accuracy and efficiency.

## II. METHODOLOGGY

*A. Pipelines Intelligent Recognition Algorithm*

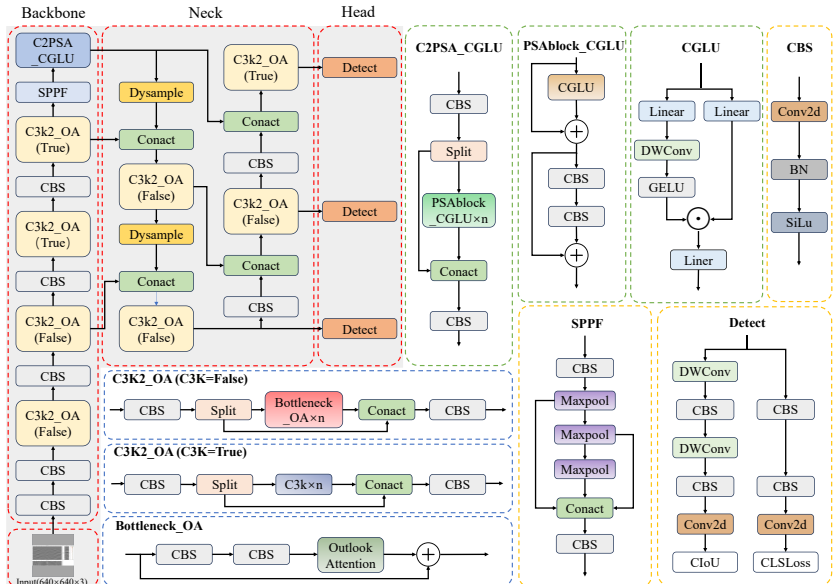

**Fig. 1** DCO-YOLO structure diagram

This study built a 3D GPR pipeline detection framework based on YOLOv11, with architectural innovations tailored to the characteristics of GPR images. YOLOv11 adopts a modular design philosophy to achieve optimization of accuracy and speed. Its core innovations are as follows:

1. Upgrading the C2f module to a C3K2 module, which integrates multi-scale convolutional kernels through a cascaded design of variable convolutional kernels to expand the receptive field and capture multi-scale spatial features.

2. Embedding the cross stage partial with pyramid squeeze attention (C2PSA) structure after the SPPF module, which utilizes a pyramid squeeze attention mechanism to enhance multi-scale feature fusion capabilities and improve target discrimination in complex backgrounds.

3. Replacing standard convolution with depth-separable convolution (DWConv), which reduces the number of parameters while maintaining classification accuracy. This study improved the C3k2, C2PSA, and Upsample modules of the YOLOv11. The improved DCO-YOLO structure is shown in **Fig. 1**.

*(1) DySample replaces nn.Upsample*

The original nn.Upsample module uses nearest-neighbor interpolation for upsampling. While efficient, it lacks dynamic adjustment capabilities, leading to loss of fine-grained details (e.g., edges of small-scale pipelines). In GPR images, pipeline edges manifest as localized geometric gradients (e.g., hyperbolic feature in B-scan). To address this, DySample, a lightweight dynamic upsampler [37], is adopted. This mechanism employs a two-stage mechanism of dynamic offset generation and Pixel Shuffle, which enables feature alignment at the sub-pixel level. It is critical for reconstructing edge features in pipeline waveforms. The offset matrix adapts to local gradients, ensuring edge continuity in upsampled feature maps (**Fig. 2**).

For the input feature map $X \in R^{H \times W \times C}$, DySample first generates the position offset $O \in R^{H \times W \times 2s^2}$ of each target sampling point directly through a lightweight linear layer, where $s$ is the upsampling factor. The dimension of the offset is $2s^2$. A dynamic adjustment mechanism is introduced to generate a dynamic range factor through a linear layer and a Sigmoid function. The equation for O is:

$$O = 0.5 sigmoid(linear_1(X)) Linear_2(X)) \qquad (1)$$

Where the function *sigmoid* is $\sigma$ in the figure, which can map any real number to the range (0, 1) and is used to control the sampling point range.

Pixel Shuffle redistributes information from the channel dimension to the spatial dimension to achieve upsampling. Offset $O$ is superimposed on the standard upsampling grid $G$ to form a dynamic sampling grid $S$. The original grid $G$ represents the fixed sampling positions of bilinear interpolation. Subsequently, the input feature $X$ is resampled according to the dynamic grid $S$ using the grid_sample function to obtain the upsampled feature $X'$.

$$X' = grid\_sample(X, S) \qquad (2)$$

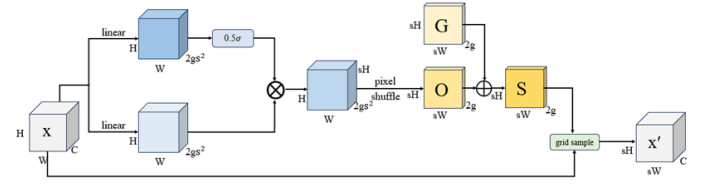

**Fig. 2** DySample structure diagram

*(2) CGLU Improves C2PSA*

The original SE module in C2PSA compresses channel-wise features into global scalars, ignoring local spatial variations. GPR images exhibit high noise heterogeneity, necessitating position-aware feature modulation. SE is replaced with CGLU (**Fig. 3**) [38], which integrates depthwise convolution into the gating branch of gated linear unit (GLU). The 3×3 depthwise convolution captures neighborhood context, generating spatially varying attention maps. This allows the model to suppress noise in low-reflectivity regions while amplifying pipeline edges. For instance, phase jumps in pipeline hyperbolas correlate with local gradient shifts. CGLU's position-dependent gating preserves these discontinuities.

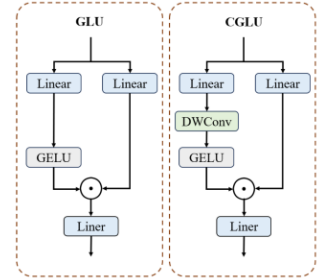

**Fig. 3** Structural diagrams of GLU and CGLU

The output of the gate branch, which captures local features, is multiplied element-wise with the output of the value branch that captures global features to generate fine-grained channel attention weights. The specific equation is as follows.

$$ConvGLU(X) = Linear_1(X) \odot GELU(DWConv(Linear_2(X))) \qquad (3)$$

Where DWConv, refers to 3×3 depth convolution, and $\odot$ represents element-wise multiplication.

CGLU not only focuses on the global features of the input image, but also observes the local features around each pixel and dynamically adjusts the channel weights at different positions. CGLU introduces 3×3 depth convolution in the gating branch to dynamically extract neighborhood features and generate position-related channel weights. In the time-frequency image of GPR signals, modulation features such as edges and phase jumps in different regions can be enhanced through this module.

*(3) OutlookAttention in C3K2*

While C3K2 leverages multi-scale kernels, it lacks explicit cross-dimensional correlation. Pipeline signatures in GPR images require joint spatial-spectral modeling. We integrate Outlook Attention into C3K2 [39]. This mechanism weights peripheral tokens relative to a central token, mimicking human visual attention. In pipeline detection, central tokens encode geometric primitives (e.g., vertex of a hyperbola), while peripheral tokens capture diffraction tails. Outlook Attention's softmax aggregation enhances coherence between these components, reducing false positives from edges. **Fig. 4** visualizes how attention weights focus on hyperbolic vertices.

The principle is shown in Fig. 4. For the input feature map $X \in R^{H \times W \times C}$, linear projections generate the value tensor $V \in R^{H \times W \times C}$ and attention weights $A \in R^{H \times W \times K^2}$ (where K is the local window size). The unfold operation extracts K×K local windows from V, reshaping it into $V_{\Delta i,j} \in R^{C \times N \times K^2}$ ($N = H \times W$ is the number of tokens). Simultaneously, a linear layer processes the center token of each window to generate $A_{i,j} \in R^{C \times N}$. Here, $A_{i,j}$ is reshaped to $R^{C \times N \times 1}$, and broadcasted to $R^{C \times N \times K^2}$ for element-wise multiplication with $V_{\Delta i,j}$:

$$Y_{\Delta i,j} = Matmul(\text{Softmax}(\hat{A}_{i,j}), V_{\Delta i,j}) \quad (4)$$

Where represents the local features after aggregation, Matmul is the matrix multiplication function, and the attention weights and values are summed up with weighting.

The Fold operation restores the aggregation results of the local window to the global feature map, as shown in the following equation.

$$\tilde{Y}_{i,j} = \sum_{0 \leq m,n \leq K} Y^{i,j}_{\Delta\ i+m-\left[\frac{K}{2}\right], j+n-\left[\frac{K}{2}\right]} \quad (5)$$

Where the input on the right side is the aggregation result $Y_{i,j} \in R^{C \times N}$ of (i, j). Sum the aggregation results of all local windows covering position (i, j) to obtain the final output $\tilde{Y}_{i,j} \in R^{H \times W \times C}$.

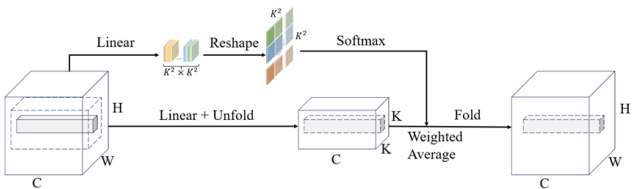

**Fig. 4** Outlook Attention structure diagram

### B. Spatial Feature Matching Method 3D-DIoU

The intersection over union (IoU) is a core metric to evaluate the performance of object detection or segmentation models. As shown in **Fig. 5**(a), it measures the localization accuracy of a model by calculating the overlap between predicted boxes and ground truth boxes. Since the three views of pipeline data have a certain spatial relationship, IoU can be used to detect the matching relationship. However, IoU is defined in a 2D plane. To detect matching relationships in 3D space, it needs to be extended to 3D-IoU. The definition of 3D-IoU is similar to IoU. The equation is as follows.

$$\text{3D IoU} = \frac{\text{Intersection Volume}}{\text{Union Volume}} = \frac{A \cap B}{A \cup B} \quad (6)$$

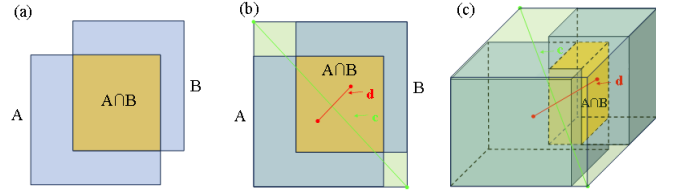

**Fig. 5** Schematic diagram of 3D-IoU (a) IoU (b) DIoU (c) 3D-DIoU

However, 3D-IoU has limitations. For model prediction, when the predicted box has no overlap with the actual box where IoU = 0, the IoU loss cannot provide gradients, resulting in the model being unable to adjust the position of the predicted box. Distance intersection over union (DIoU) is a loss function used for bounding box regression in object detection [40]. It introduces a center point distance penalty term, improving the convergence speed and localization accuracy of detection models, as shown in **Fig. 5** (b). When the two bounding boxes are far apart, the penalty term of DIoU increases with the center point distance, better aligning with the logical judgment of box positions in real-world scenarios. Extending DIoU to three-dimensional space, as shown in **Fig. 5**(c), yields the 3D-DIoU for two three-dimensional targets. The equation for calculating 3D-DIoU for three-dimensional targets is as follows.

$$\text{3D DIoU} = \text{3D IoU} - \frac{d^2}{c^2} \quad (7)$$

Where $d$ is the Euclidean distance between the centers of the two boxes. $c$ is the diagonal length of the smallest cuboid that covers the two boxes.

### III. GPR PIPELINE DATA PROCESSING AND FAST IDENTIFICATION METHODS

#### A. Pipeline Three-View Image Feature Analysis

3D GPR constructs volumetric data through the fusion of orthogonal scanning data, with imaging interpretation relying on three fundamental geometric perspectives: B-scan, C-scan, and D-scan, whose spatial relationships are illustrated in **Fig. 6**.

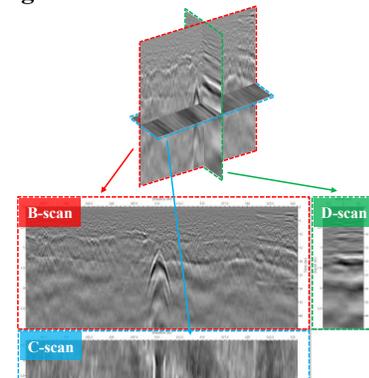

**Fig. 6** 3D view of the spatial relationship of 3D GPR

To analyze the three-view imaging characteristics of different pipeline types, this study employs FDTD simulations for forward modeling. A three-layer road model comprising a 20cm surface course, 40cm base course and 65cm subgrade course was established based on actual road structures and 3D GPR detection parameters. All structural layers adopted ideal contact boundary conditions, with a fixed transmitter antenna height of 0.05 m. Critical model parameters are summarized in Table I.

TABLE I
FORWARD SIMULATION PARAMETERS

| Setting parameters | Preset value |
|---|---|
| Forward modeling model size (m) | 6.0 × 2.0 × 3.0 |
| Spatial grid step (m) | 0.005× 0.005× 0.005 |
| Time window (ns) | 60 |
| Antenna travel distance (m) | 0.1 |
| Antenna type / frequency | ricker / 600MHz |
| Number of lines | 8 |
| Line spacing (m) | 0.25 |
| Air material ($\varepsilon_r / \sigma / \mu_r / \sigma_*$) | 1 / 0 / 1 / 0 |
| Surface material ($\varepsilon_r / \sigma / \mu_r / \sigma_*$) | 4 / 0.001 / 1 / 0 |
| Base material ($\varepsilon_r / \sigma / \mu_r / \sigma_*$) | 7 / 0.001 / 1 / 0 |
| Subbase material ($\varepsilon_r / \sigma / \mu_r / \sigma_*$) | 7 / 0.001 / 1 / 0 |
| Metallic material ($\varepsilon_r / \sigma / \mu_r / \sigma_*$) | 1 / 7×10$^6$ / 100 / 0.4 |

This study established three typical pipeline models: vertical pipelines, horizontally inclined pipelines and deeply inclined pipelines. All pipelines were made of metal materials with a diameter of 0.5 m, a wall thickness of 5 mm and filled with air. The specific geometric configurations are shown in **Fig. 7**. Vertical pipelines were distributed perpendicular to the x-axis, with their center coordinates set at (3.0, 1.3, 0.0) and (3.0, 1.3, 2.0). Horizontal inclined pipelines were arranged at an angle in the x-z plane, with their centers located at (2.7, 1.3, 0.0) and (3.3, 1.3, 2.0), forming a sloped straight-line structure at a 16.7° angle with the x-axis. The deeply inclined pipelines exhibited a increasing burial depth within the y-z plane, with the center coordinates of the two ends designed as (3.0, 1.1, 0.0) and (3.0, 1.5, 2.0), forming inclined pipelines at an angle of 11.3° to the y-axis. The above parameter settings provide a reliable numerical simulation foundation for subsequent algorithm verification.

Forward simulation results are shown in **Fig. 8-10**, while field measurement results are presented in **Fig. 11**. Comparative analysis of 3D GPR images reveals high consistency between simulated and measured data in reflection signature morphology, geometric correspondence, and phase continuity, validating the authenticity and reliability of the experimental GPR dataset.

In B-scan, all three pipeline types exhibit characteristic hyperbolic reflection signatures. The vertex coordinates ($x_0$, $d$) of these hyperbolas relate to pipeline burial depth $d$.

$$d = \frac{v}{2}\sqrt{(\frac{t_0}{2})^2 - (\frac{x_0}{v})^2} \quad (8)$$

Where $v$ is the electromagnetic wave velocity and $t_0$ is the arrival time at the hyperbolic vertex. While B-scan images alone cannot effectively distinguish pipeline spatial orientations, they clearly display diffraction waves at pipe ends and multiple reflections from pipe bottoms. Consequently, all three pipeline types are uniformly annotated as 1-B in B-scan images.

C-scan demonstrate distinct differences across pipeline types. Vertical pipelines exhibit two parallel linear features aligned with the survey direction, annotated as 1-C. Horizontal inclined pipelines display parallel linear structures with angular offsets, where the included angle reflects the pipeline's horizontal inclination, annotated as 2-C. Deeply inclined pipelines show non-parallel linear features with scale differences between upper and lower edges, annotated as 3-C.

D-scan results further differentiate pipeline orientations. Vertical pipelines present horizontal layered reflections, annotated as 1-D. Horizontal inclined pipelines share hyperbolic features, annotated as 2-D. Deeply inclined pipelines reveal tilted parallel reflection whose inclination direction corresponds to the pipeline's depth-wise direction, annotated as 3-D.

B-scan images could not distinguish spatial orientations of pipelines. In contrast, C-scan and D-scan views show orientation-specific features. Thus, multi-view fusion is essential for accurate 3D localization. The FDTD-based simulated dataset serves exclusively as a validation tool for multi-view feature consistency analysis. This approach ensures controlled-variable verification of pipeline signatures across three views (B/C/D-scan) while avoiding computational constraints associated with large-scale simulations.

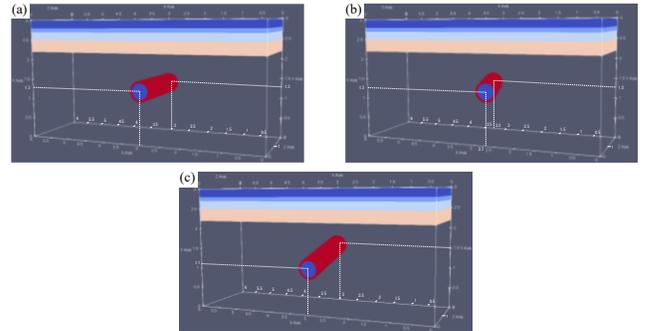

**Fig. 7** Forward pipeline model (a) Vertical pipeline (b) Horizontal inclined pipeline (c) Deeply inclined pipeline

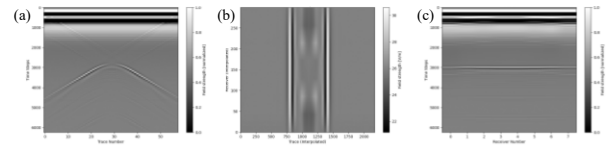

**Fig. 8** Forward modeling results for vertical pipeline (a) B-scan (b) C-scan (c) D-scan

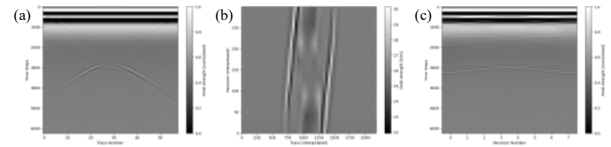

**Fig. 9** Forward modeling results of horizontal inclined pipeline (a) B-scan (b) C-scan (c) D-scan

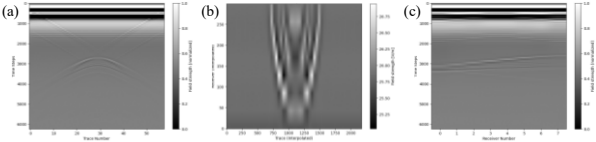

**Fig. 10** Forward modeling results of deeply inclined pipeline (a) B-scan (b) C-scan (c) D-scan

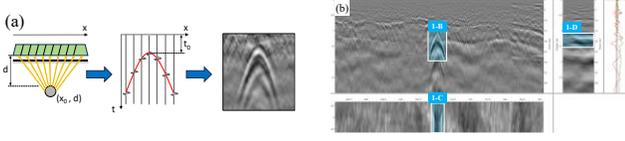

**Fig. 11** Actual image characteristics of GPR pipeline (a) Pipeline B-Scan image imaging principle (b) Vertical pipeline three-view (c) Horizontal inclined pipeline three-view (d) Deeply inclined pipeline three-view

*B. Dataset Construction*

The 3D GPR dataset construction process proposed in this study is shown in **Fig. 12**, covering the four links of data acquisition, data preprocessing, image processing and dataset establishment.

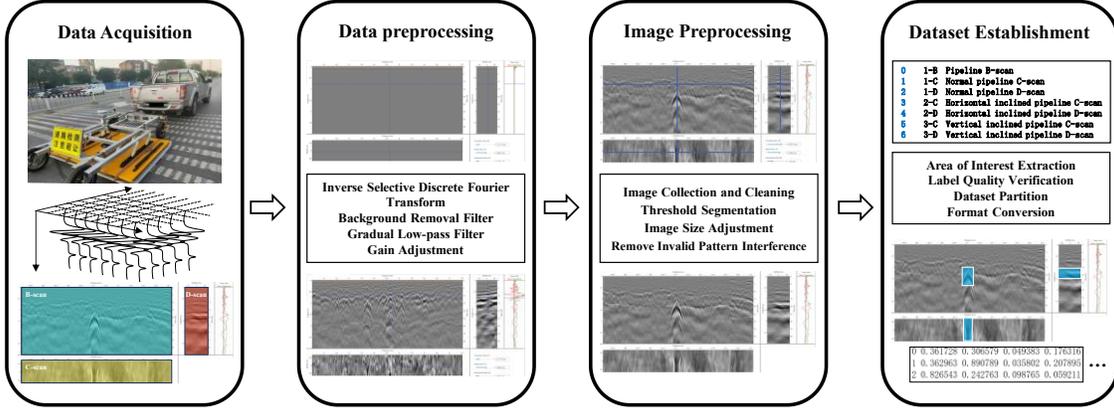

**Fig. 12** Dataset construction process

(1) Data Acquisition

The 3D GPR system was used for data acquisition, which consists of the GeoScope$^{TM}$ MKIV radar mainframe, the 8-channel ground-coupled antenna DXV1808 with a frequency range of 90 MHz - 1.0 GHz, the real-time kinematic high-precision localization device and the distance measurement indicator. The antenna is based on stepped-frequency continuous wave technology (SFCW), which can meet shallow high-resolution and deep structure detection. The main data acquisition parameters are as follows: the trigger spacing is 90mm, the sampling rate is 1024, the dwell time is 2us, and the time window is 125ns.

The measured area was selected from the main roads of Shenyang, China, encompassing a total survey mileage of 100 km. The surveyed environments included areas with different pavement structures and traffic loading conditions. There is currently no publicly available large-scale 3D GPR dataset specifically dedicated to underground pipeline detection with three-view images. The dataset serves as a significant contribution in this field. All data originated from this collected field measurement campaign.

While the FDTD-simulated data provides valuable insights for qualitative multi-view feature analysis, it was excluded from the training pipeline. The significant gap between even the most sophisticated simulations and real GPR data. Key discrepancies include: (1) The inability of forward models to completely replicate the complex noise patterns prevalent in field data. (2) Idealized assumptions in simulations regarding material homogeneity and geometry, which contrast with the unpredictability and heterogeneity of real underground environments.

Incorporating simulated data for augmentation risks biasing the model towards learning these idealized features, potentially degrading its performance and generalization capability when deployed on real-world data [3, 41]. Thus, priority was given to learning from a large volume of authentic, annotated field measurements. The high performance achieved on the held-out real-world validation set confirms the sufficiency of this data-centric strategy.

*(2) Data Preprocessing*

For the frequency domain signal characteristics of SFCW GPR, the frequency domain signal is converted to time domain signal by inverse selective discrete Fourier transform (ISDFT) to optimize the time domain resolution. The equation is as follows:

$$s(t) = \sum_{k=1}^{N} S(f_k) e^{j2\pi f_k t} \quad (9)$$

Where $S(f_k)$ is the k$^{th}$ frequency component, $N$ is the number of frequency sampling points, $s(t)$ is the time domain signal.

In addition, the interference signals caused by antenna direct-coupled waves, surface reflections and system noise were eliminated by the background removal filter. Through the gradual low-pass filter to suppress high-frequency electronic noise and low-frequency surface reflection interference, eliminate periodic noise and smooth random noise. Compensate for signal attenuation effect by exponential gain adjustment to enhance the visibility of deep targets.

To validate the contribution of preprocessing steps, an ablation experiment was designed (Table II). Information entropy (*IE*) was employed as the metric, which quantifies the information richness of a preprocessed GPR image as follows:

$$IE = -\sum_{i=0}^{L-1} p(i)\log_2(p(i)) \qquad (10)$$

Where *IE* is the image information entropy; *L* is the total number of grayscale levels; *p(i)* denotes the ratio of pixels with grayscale value *i* to the total number of pixels in the image.

Higher *IE* values indicate superior preservation of discriminative pipeline features like edge gradients, hyperbolic signatures. Each configuration incrementally removes one step while retaining ISDFT as the foundational transformation. All steps served as the baseline. The observed degradation in *IE* upon the individual exclusion of each preprocessing step substantiates the effectiveness of all methods in improving signal-to-noise ratio of GPR data.

Table II
GAIN ABLATION EXPERIMENT OF DATA PREPROCESSING

| ISDFT | Gain adjustment | Background removal filter | Gradual low-pass filter | *IE* | *ΔIE* |
|---|---|---|---|---|---|
| √ | √ | √ | √ | 3.6757 | baseline |
| √ | √ |   | √ | 3.1620 | ↓14.0% |
| √ | √ | √ |   | 3.5757 | ↓2.7% |
| √ | √ |   |   | 3.0876 | ↓16.0% |
| √ |   |   |   | 1.3601 | ↓63.0% |

*(3) Image Processing*

The 3D GPR data were decomposed to obtain the B/C/D-scan images. The image parameters were set to a depth of 3 m, a span of 30 m and a vertical span direction of 1.32 m. The image size was adjusted and the contrast was optimized through standardized processing, and the irrelevant patterns and text annotations were removed.

*(4) Dataset Establishment*

The three-view GPR images were labeled using LabelImg software. The number of labeled boxes is shown in Table III. To ensure the annotation consistency, a rigorous, multi-stage labeling protocol was established and followed. Firstly, detailed annotation guidelines were created, defining precise criteria for bounding box placement for each pipeline class across all three views. For example, in B-scan images, boxes were required to enclose the entire hyperbolic reflection. Secondly, all annotators underwent comprehensive training using these guidelines and a common set of reference images to calibrate their understanding. Finally, a senior researcher, who is an author of this paper, reviewed over 30% of the annotations randomly selected from each batch. Any ambiguous cases were discussed and resolved through consensus, ensuring label integrity across the entire dataset. The dataset was divided into training set and validation set according to the ratio of 8:2. The labeled files were converted to .txt format.

Table III
DATASET AND NUMBER OF LABELED BOXES

| Category | Training set | Validation set |
|---|---|---|
| Image | 1560 | 391 |
| 1-B | 2046 | 505 |
| 1-C | 1468 | 380 |
| 1-D | 1249 | 339 |
| 2-C | 266 | 26 |
| 2-D | 164 | 39 |
| 3-C | 175 | 31 |
| 3-D | 150 | 37 |

*C. Pipeline Image Characterization and Performance*

The experimental hardware environment adopted a high-performance computing platform constructed by NVIDIA RTX 3060 GPU with 12GB memory. The model training and validation were based on the PyTorch. The Adam optimizer was used for network optimization; the initial learning rate was set to 0.01 and the batch size was set to 32. The maximum number of iteration rounds was set to 140 epochs during the training process, and the loss of the validation set was monitored by the early stopping mechanism with a patience of 20. The training was terminated early in 118 epochs when a better loss did not appear in 20 consecutive epochs, so that the final number of training rounds was 120 epochs.

In this paper, average precision (AP), precision (P), recall (R), mean average precision (mAP), F1 value and other indicators were used to evaluate the detection effect of the model.

P and R are two related quantities, P in the process of increasing, each point will correspond to a value of R. The area enclosed by the curve and the coordinate axes from the combination of these points is AP, regarding as the average of the precision values corresponding to different recalls. mAP represents the average accuracy. The F1 value is the reconciled average of the accuracy and recall, and is intended to reflect both the model's recognition accuracy and coverage of positive class samples.

*D. Pipeline Rapid Determination Methods*

While 3D GPR can characterize pipeline spatial features through B/C/D-scan views, efficiently correlating cross-view features remains challenging in practical applications. Current methods for pipeline spatial feature analysis based on 3D GPR primarily rely on manual interpretation and direct 3D data volume processing. Manual interpretation is time-consuming and prone to subjective errors. Other 3D data processing methods such as 3D-CNN can model spatial continuity. But they suffer from exponentially growing computational complexity caused by cubic convolution kernel traversal through high-dimensional data. Therefore, this study proposes a lightweight 3D mapping mechanism that significantly reduces computational load by efficiently mapping 2D features from three views into 3D space. Specifically, based on the 3D coordinate system constructed from three-view frames of 3D GPR data, the method extends 2D annotation boxes from B/C/D-scan views to 3D bounding boxes. A multi-view matching algorithm based on 3D-DIoU automatically selects spatially consistent candidate combinations, enabling rapid classification of 3D pipeline.

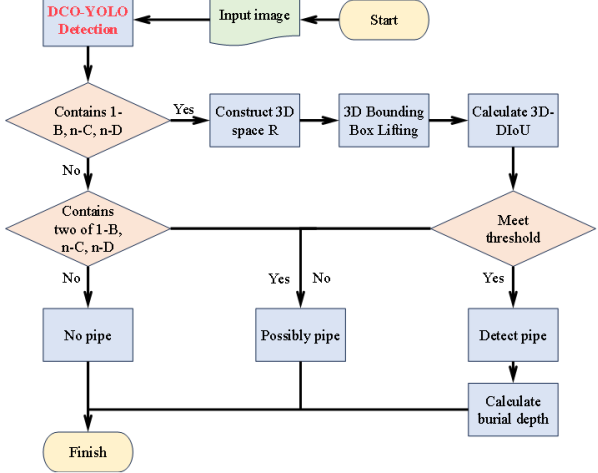

**Fig. 13** Pipeline spatial position determination structure

The rapid pipeline spatial positioning discrimination framework is illustrated in **Fig. 13**. Pipeline directions are categorized into three types: (1-B, 1-C, 1-D) represents vertical pipelines, (1-B, 2-C, 2-D) represents horizontal inclined pipelines, and (1-B, 3-C, 3-D) represents deeply inclined pipelines. The existence of a pipeline requires not only detecting all features in any group but also verifying their spatial consistency. To achieve this, the three-view boxes from the input images are reconstructed into the 3D coordinate system R, building a detection model with full spatial dimensionality. Specifically, 1-B provides pipeline cross-sectional width along the x-axis and burial along the z-axis; n-C contains planar trajectory features across x and y axes; while n-D carries spatial coordinates reflecting depth-direction inclination across y and z axes. Through 3D spatial mapping, orthogonal view geometric features are innovatively introduced to compensate for missing dimensions: integrating n-C's y-axis coordinates into 1-B forms 3D-1B, compensating n-D with 1-B's z-axis parameters generates 3D-nD and enhancing n-C with n-D's x-axis features constructs 3D-nC. Taking 1-B's x-axis as an example:

$$x_{B1} = \frac{X_{B1} - main\_view(X)}{main\_view(width)} \quad (11)$$

$$x_{B2} = \frac{X_{B2} - main\_view(X)}{main\_view(width)} \quad (12)$$

Where $x_{B1}$, $x_{B2}$ represent the coordinate values of the left and right edges of the 1-B labeling frame converted to the space R. $X_{B1}$, $X_{B2}$ represent the coordinate values of the left and right edges of the 1-B labeling frame under the input image, respectively. Main_view deposits the coordinates and dimensions of the upper left vertex of the main view.

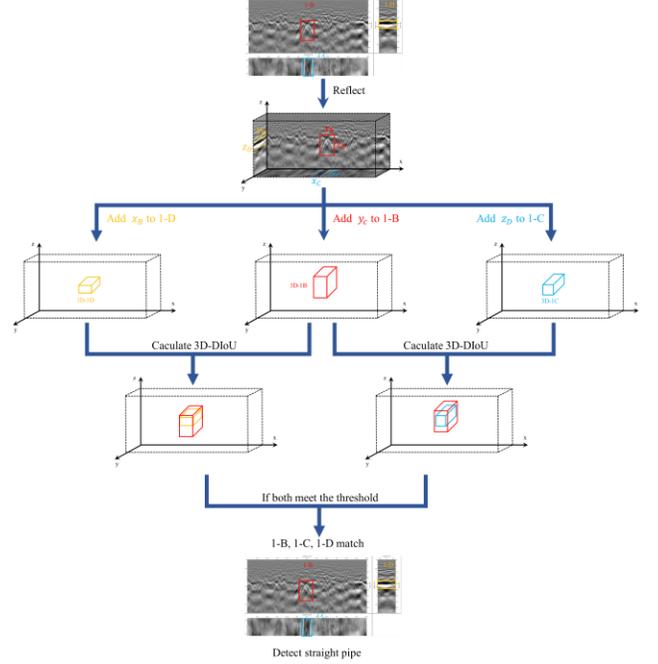

**Fig. 14** Pipeline feature matching discrimination process

Under ideal linear projection conditions, reconstructed 3D-1B, 3D-nC and 3D-nD should perfectly coincide. However, multiple interference factors in 3D GPR data challenge this assumption, including: Fresnel effect-induced scattering feature distortion, spatial reference deviations from multi-view data acquisition intervals, and scale normalization differences in preprocessing algorithms. A 3D-DIoU assessment system based on normalized spatial parameters is proposed to evaluate the spatial similarity among 3D-1B, 3D-nC, and 3D-nD, as shown in **Fig. 14.** First, three view boxes are mapped into 3D space. After feature completion, three 3D objects 3D-1B, 3D-nC, 3D-nD in the same coordinate system undergo pairwise 3D-DIoU calculations using Equation 7 to determine view feature matching through threshold comparison.

The 3D-DIoU threshold critically determines spatial matching accuracy. To determine this threshold, statistical analysis was conducted on the 3D-DIoU distribution across all annotated samples **(Fig. 15)**. Results showed that 91.7% of samples had B-scan/D-scan 3D-DIoU above 0.4, with all samples demonstrating B-scan/C-scan 3D-DIoU exceeding 0.4. Setting the 3D-DIoU matching threshold below 0.4 effectively reduces false detection rates. To emulate environmental heterogeneity, additive Gaussian noise was injected into the original 3D GPR images across all three views. The noise intensity was governed by the standard deviation ($\sigma$), with two levels evaluated. Under the medium-noise condition ($\sigma = 0.1$), partial characteristics remain discernible, whereas under the high-noise condition ($\sigma = 0.2$), only approximate outlines are perceptible. As summarized in Table IV, the proportion of data with 3D-DIoU values above 0.4 consistently exceeds 90%. These results demonstrate that a 3D-DIoU threshold of 0.4 maintains high selection accuracy, even under noisy conditions.

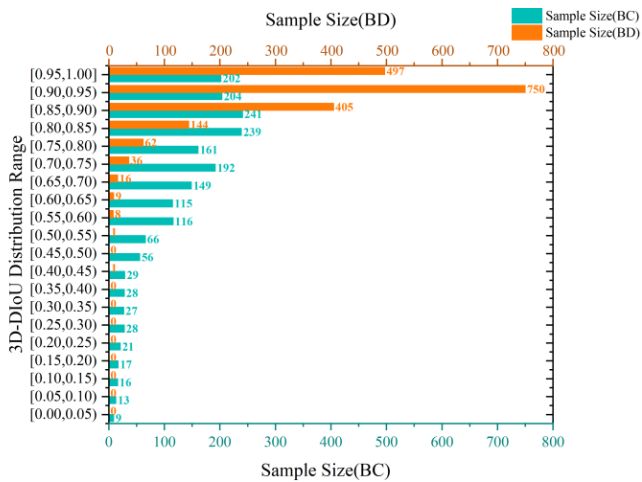

**Fig.15** Distribution map of 3D-DIoU of matching pipeline

Table IV
3D DIoU STATISTICAL RESULTS

| Condition | B-Scan & C-Scan (<0.4) | B-Scan & C-Scan (>0.4) | B-Scan & D-Scan (<0.4) | B-Scan & D-Scan (>0.4) |
|---|---|---|---|---|
| Original | 0 | 100.00% | 8.2% | 91.8% |
| $\sigma = 0.1$ | 0.05% | 99.95% | 7.9% | 92.1% |
| $\sigma = 0.2$ | 0.60% | 99.4% | 5.8% | 94.2% |

To validate the algorithm's lightweight characteristics, we conducted data volume comparative analysis on a 1 km road section with standard 3.5 m lane width. Using standard RGB image inputs of 1620 × 760 pixels in JPG format at approximately 250 KB per image, each image corresponds to 30 m GPR data per channel. For 35 channels covering the road width, the 1 km detection task generates 1,200 images totaling 300 MB. In contrast, 3D data volume processing requires sampling points every 5 cm along the road length equivalent to 20,000 points per kilometer, with 2,048 depth points per channel at 4 bit per point, totaling 20,000 × 35 × 2,048 × 4 bit = 5,470 MB. Results demonstrate that our algorithm reduces input data volume to 5.6% of conventional 3D processing methods while maintaining spatial feature accuracy, significantly lowering storage and computational demands.

## IV. RESULTS AND DISCUSSION

### A. Model Performance Evaluation

The training performance evolution of the DCO-YOLO pipeline recognition model is shown in **Fig. 16**. The mAP@50 and box_loss metrics exhibit a typical three-stage learning curve during training: an initial rapid changes phase spanning 0-20 epochs, a gradual convergence phase spanning 20-80 epochs, and a stable saturation phase spanning 80-120 epochs.

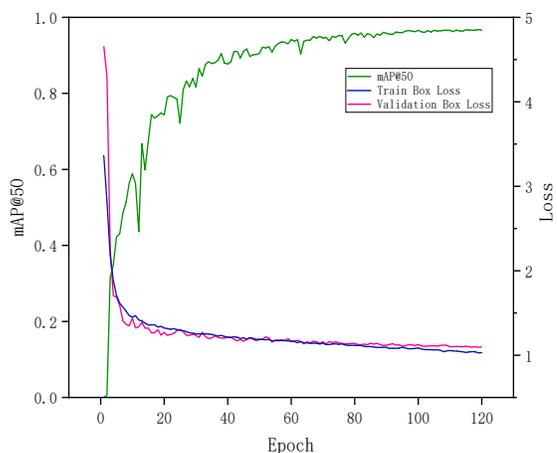

**Fig. 16** DCO-YOLO mAP@50 and Loss curves

During the 0-20 epoch phase, mAP@50 achieves exponential growth, with the total loss decreasing by over 90%; in the 20-80 epoch phase, the growth rate of mAP@50 decreases to 0.28% per epoch, effectively avoiding local optima traps; after 80 epochs, the model enters a convergence state, with the fluctuation range of mAP@50 narrowing to ±1%. Box_loss stabilizing below 1.05, indicating that the network parameters have reached an optimal configuration.

The P-R curve quantitatively characterizes the precision-recall trade-off relationship of the classification model under the dynamic threshold, and its area under the curve (AP) is a key index to evaluate the comprehensive performance of the model. As shown in **Fig. 17**(a), the P-R curves of the seven categories of target detection tasks show some differences: the AP values of categories 3-C and 3-D are 0.931 and 0.907, respectively, which are lower than those of the other categories with an average AP of 0.967. In contrast, the P-R curves for categories 1-B and 1-C converge to the upper-right quadrant of the coordinate axis, with AP values exceeding 0.97, indicating that the model has excellent precision-recall balance for vertical pipeline detection.

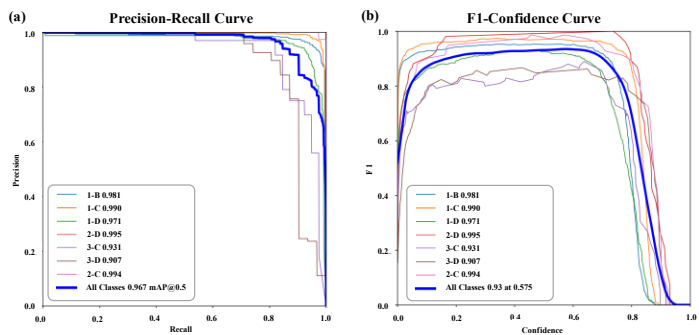

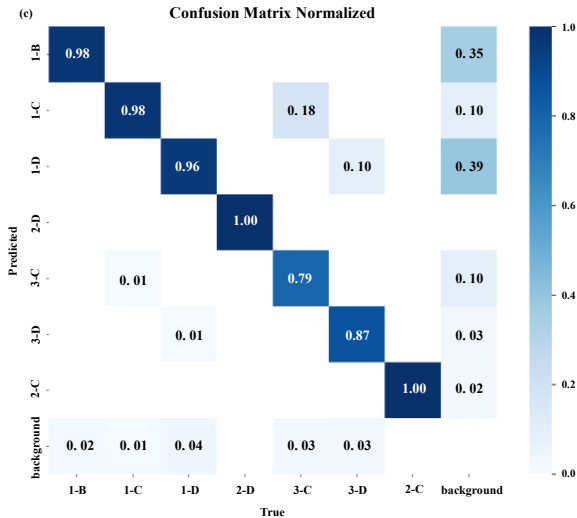

**Fig. 17** DCO-YOLO recognition results (a) P-R curve (b) F1 curve (c) confusion matrix

The F1 confidence curve reveals the performance optimization path of the model under different confidence thresholds. As shown in **Fig. 17**(b), the weighted average F1 value of the model exceeds 0.9 when the confidence threshold is in the range of the interval 0.3-0.7. In particular, the weighted average F1 value of the model reaches a peak value of 0.93 when the threshold is set at 0.575. This experimental result provides a basis for optimizing key parameters for engineering applications.

The confusion matrix in **Fig. 17**(c) provides a class-wise performance diagnosis, which is particularly valuable for assessing the model's capability on the imbalanced classes (2-C, 2-D, 3-C, 3-D). While the overall accuracy for these inclined pipeline categories is commendable, the matrix reveals specific misclassification patterns. For instance, 18% of 3-C samples were misclassified as the 1-C. This occurs because the imaging features of deeply inclined pipelines can, from certain perspectives, resemble those of vertical pipelines, especially when the inclination angle is subtle.

The critical question raised by the class imbalance is whether these rare categories are still robustly detected. The P-R curves in **Fig. 17**(a) offer a positive answer. The AP values for the rare classes (e.g., 0.931 for 3-C, 0.907 for 3-D), while lower than those for the abundant classes, remain above 0.90. This indicates that when the model does make a prediction for these classes, it is highly reliable. This performance is achieved without employing explicit countermeasures for class imbalance. These techniques were avoided for two reasons: First, to prevent the potential introduction of overfitting on the rare classes or the generation of unrealistic synthetic data; Second, the performance on these rare classes was deemed sufficient for the practical application scenario, where the primary goal is the accurate identification and spatial typing of pipelines. The high recall for all classes, as seen in the confusion matrix where the true positives are dominant, confirms that the model effectively avoids false negatives, which is the most critical requirement in pipeline safety inspection.

*B. Ablation Experiment*

The ablation experiments were conducted to quantitatively evaluate the contribution of each improved module to the model performance. The results are summarized in Table V. The original YOLOv11 algorithm achieved precision of 94.2%, recall of 91.2%, mAP@50 of 95.8% and mAP@50-95 of 67.7%. The integration of the DySample, CGLU and OutlookAttention modules significantly improved the algorithm's performance. Ablation experiments show that compared to the original YOLOv11, DySample introduces a dynamic point sampling mechanism that effectively preserves B-scan hyperbolic edge features. The CGLU module dynamically extracts neighborhood features and generates position-related channel weights. This suppressed background noise while focusing on local features, improving precision by 0.6%, recall by 0.5%. The OutlookAttention module enhances the model's ability to capture local fine-grained features, improving precision by 1% compared to the single DySample module. The improved DCO-YOLO achieves a 2% improvement in precision, a 2.1% improvement in recall, a 0.9% improvement in mAP@50, and a 3.5% improvement in mAP@50-95 compared to YOLOv11, validating the effectiveness of the multi-dimensional feature enhancement strategy.

Table V
MODEL ABLATION EXPERIMENT

| DySample | CGLU | OutlookAttention | Precision (%) | Recall (%) | mAP@50 (%) | mAP@50-95 (%) |
|---|---|---|---|---|---|---|
|  |  |  | 94.2 | 91.2 | 95.8 | 67.7 |
| √ |  |  | 94.7(+0.5) | 91.8(+0.6) | 96.1(+0.3) | 68.4(+0.7) |
| √ | √ |  | 95.3(+1.1) | 92.3(+1.1) | 96.3(+0.5) | 68.8(+1.1) |
| √ |  | √ | 95.7(+1.5) | 91.9(+0.7) | 96.2(+0.4) | 69.5(+1.8) |
| √ | √ | √ | **96.2(+2.0)** | **93.3(+2.1)** | **96.7(+0.9)** | **71.1(+3.5)** |

To further investigate the interaction among the three modules and validate their co-optimization, a progressive attention analysis using Grad-CAM++ was conducted across three model variants. Compared with Grad-CAM, Grad-CAM++ introduces second-order gradient calculation and weighted gradient mechanisms to better capture the distribution of multiple similar targets in an image. As depicted in **Fig. 18**, the heatmaps reveal a complementary evolutionary path.

The baseline YOLOv11 exhibits diffuse and scattered attention regions, often extending beyond the pipeline contours. The introduction of DySample dynamically enhances the model's focus on the spatial extent of pipeline edges. However, these attention regions still exhibit fluctuations and lack of concentration. The integration of the CGLU and OutlookAttention module addresses this limitation. The CGLU's position-aware gating mechanism and the attention concentration of OutlookAttention effectively suppresses the erratic activations. Furthermore, even when there are many pipeline waveforms, the red regions still accurately cover the contours of most waveforms. These observations confirm the improved model's dual advantages regarding feature localization accuracy and robustness.

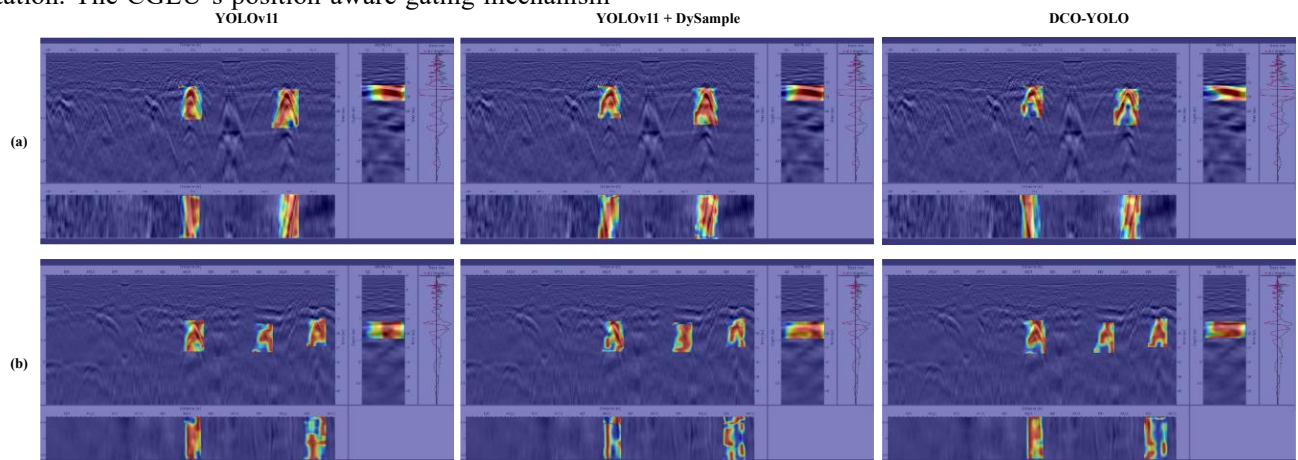

**Fig. 18** Comparison of heat map results for YOLOv11, YOLOv11+DySample and DCO-YOLO

*C. Model Comparison*

To evaluate the effectiveness of the proposed algorithm, comprehensive comparative experiments were conducted. The tested algorithms include Faster R-CNN, SSD, several YOLO series models and DCO-YOLO. All models were trained on the same dataset using the same parameter settings for 120 epochs.

Table VI summarizes the results of these comparative experiments. DCO-YOLO achieved the highest scores of 96.2% precision, 93.3% recall, and 96.7% mAP@50, while simultaneously demonstrating inference speeds of 75.88 frames per second (FPS) on NVIDIA RTX 3060, outperforming all baseline models in computational efficiency. These results highlight the effectiveness of the model modifications. The mAP@50 comparison indicates that single-stage algorithms exhibit faster convergence than two-stage algorithms. With the frame rate advantage further validating their suitability for resource-constrained deployment scenarios. Over 50 FPS is critical for mobile 3D GPR field detection systems. However, end-to-end frame rates and power consumption are subject to additional variables including GPR data acquisition latency and sensor synchronization overhead. These factors fall beyond the scope of our algorithmic contribution, which focuses on computational efficiency in the image analysis pipeline.

Table VI
MODEL COMPARISON TEST

| Model | Precision (%) | Recall (%) | mAP@50 (%) | FPS |
|---|---|---|---|---|
| Faster R-CNN | 73.2 | 84.9 | 84.6 | 6.88 |
| SSD | 92.4 | 87.0 | 93.7 | 56.41 |
| YOLOv5 | 92.3 | 89.0 | 95.2 | 64.93 |
| YOLOv8 | 92.2 | 92.0 | 95.4 | 72.65 |
| YOLOv10 | 88.4 | 86.7 | 93.6 | 63.61 |
| YOLOv11 | 94.2 | 91.2 | 95.8 | 65.34 |
| DCO-YOLO | **96.2** | **93.3** | **96.7** | **75.88** |

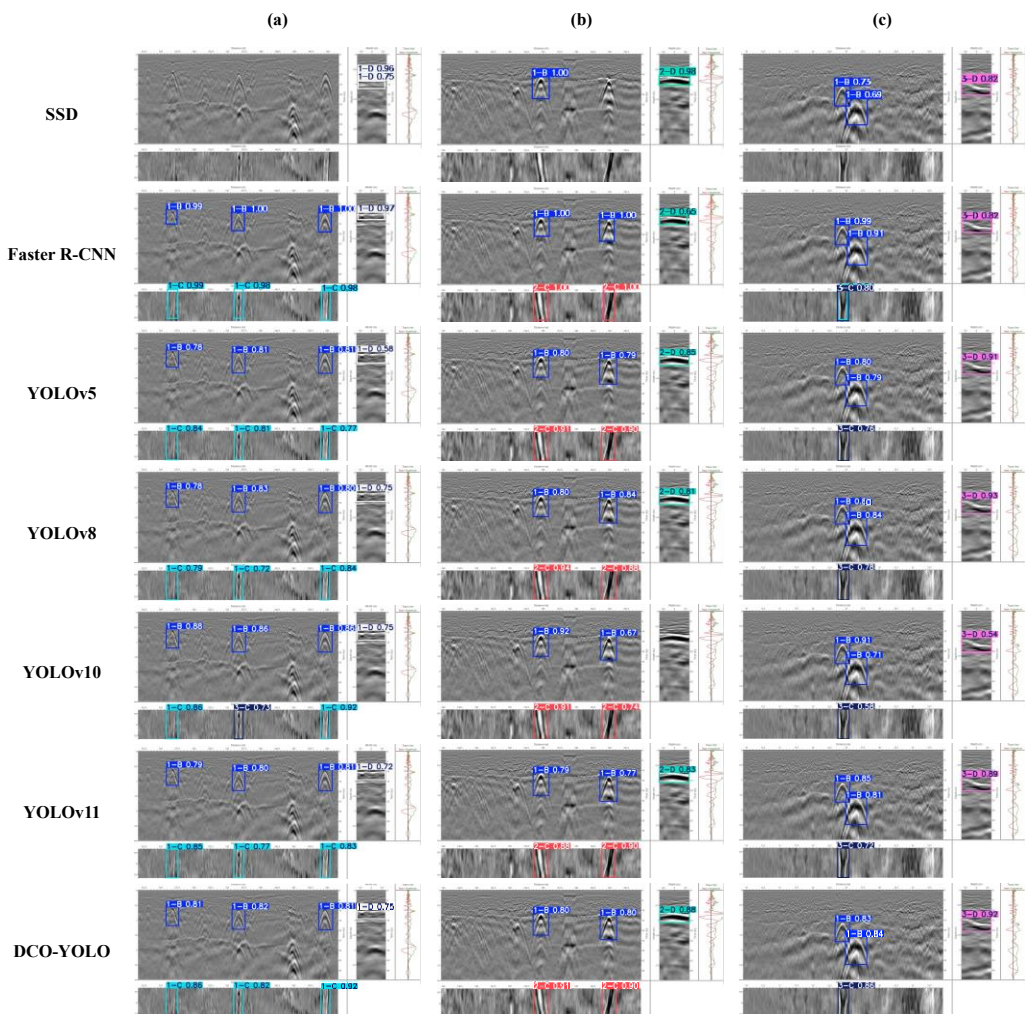

**Fig. 19** Model comparison test results

**Fig. 19** provides an intuitive comparison of the detection performance of different algorithms. Although Faster R-CNN achieves higher bounding box confidence scores, it has a higher false positive rate compared to YOLO-based models. Furthermore, DCO-YOLO achieves the highest detection accuracy for various hidden defect types, highlighting the impact of its targeted model improvements on interpreting complex GPR features.

*D. Pipeline Rapid Identification System*

This study developed a pipeline rapid classification system based on the above research findings, as shown in **Fig. 20**. The prediction confidence, prediction IoU, and matching 3D-DIoU thresholds are all adjustable, with default values of 0.5, 0.7 and 0.4. In practical applications, these thresholds can be adjusted according to the actual situation. For example, when there is a high level of noise, the prediction threshold can be increased. When pipelines are densely distributed, the matching threshold can be increased. The pipeline discrimination system can accurately identify pipeline features and achieve matching of different view features of the same pipeline and depth estimation. Additionally, the system supports batch processing of GPR images, enabling rapid discrimination of many input images, significantly improving the processing efficiency of 3D GPR data.

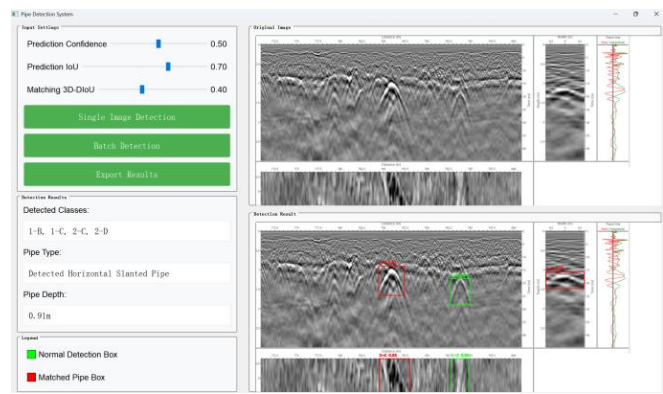

**Fig. 20** Pipeline rapid identification system

V. CONCLUSION

This study addresses the issues of weak correlation between multi-view features, low recognition accuracy of small-scale targets, and insufficient robustness in complex scenarios in 3D GPR underground pipeline detection. A multi-strategy improved deep learning framework is proposed for intelligent 3D pipeline detection. The key conclusions are as follows:

1. A pipeline identification and localization method based on the joint analysis of B/C/D-scan images is established. Through FDTD forward simulation and validation with extensive real-world measurement data, it is demonstrated that multi-view joint analysis significantly enhances the accuracy of pipeline orientation identification. This approach effectively overcomes the inherent ambiguity of single-view detection and the high computational burden of direct 3D volume processing.

2. By innovatively improving the YOLOv11 architecture — integrating DySample dynamic upsampling for fine-grained edge reconstruction, CGLU for position local feature perception, and the OutlookAttention mechanism for cross-dimensional correlation—the model's performance is substantially enhanced. Experimental results confirm that the proposed DCO-YOLO achieves precision, recall, and mAP@50 of 96.2%, 93.3%, and 96.7%. Grad-CAM++ visualizations further validate that DCO-YOLO focuses more precisely on the local geometric details and edge contours of pipeline signatures.

3. The proposed 3D-DIoU spatial feature matching algorithm enables automated association of pipeline annotations across different views. This provides a theoretical foundation for developing rapid, on-site pipeline identification systems.

While the proposed framework demonstrates superior performance on the tested urban pipeline dataset, several limitations and opportunities for future research warrant discussion:

(1) The current model is validated primarily on data collected with a specific 3D GPR system in typical urban road environments. The gap between different GPR systems and survey environments poses a challenge for generalization. (2) The efficacy of the 3D-DIoU matching algorithm relies on the accurate detection and geometric consistency of features across three views. (3) Although the multi-view 2D approach is significantly more efficient than processing full 3D volumes, the DCO-YOLO model and the subsequent matching logic still require GPU acceleration for real-time analysis. Further optimization and lightweight deployment are necessary.

Future work will focus on advancing intelligent underground pipeline detection towards greater robustness and practicality. Exploring domain adaptation and meta-learning techniques to improve model generalization across different GPR systems and heterogeneous geological environments. Investigating robust fusion algorithms that can tolerate missing or corrupted views. Developing lightweight model variants suitable for real-time processing on embedded systems. Integrating multi-modal sensor data as LiDAR, InSAR to achieve collaborative surface-subsurface perception, thereby providing a more comprehensive solution for smart city underground infrastructure management.

ACKNOWLEDGMENT

This research was funded by the National Key Research and Development Program of China [grant number 2023YFB2603500]; Project of Shenzhen Science and Technology Plan [grant number KJZD20230923115206014]; Heilongjiang Natural Science Foundation Research Team Project [grant number TD2022E001].

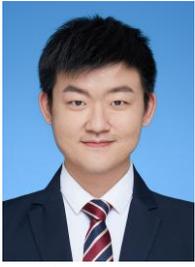
**Haotian Lv** received the B.S. degree in Civil Engineering from Shandong University, Jinan, China, in 2020. He has been a Ph.D. candidate in the School of Transportation Science and Engineering, Harbin Institute of Technology, Harbin, China since 2020. His research interests include non-destructive testing, ground penetrating radar (GPR) and deep learning (DL) methods in GPR image recognition.

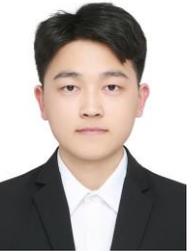
**Chao Li** is currently pursuing the B.S. degree in Traffic equipment and control engineering in the School of Transportation Science and Engineering, Harbin Institute of Technology, Harbin, China. His research focuses on 3D ground penetrating radar (GPR) signal processing and intelligent interpretation systems, with particular interests in subsurface object detection using deep learning architectures.

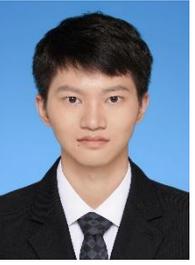
**Jiangbo Dai** received the B. S. degree in Road and Bridge Engineering from Harbin Institute of Technology, Harbin, China, in 2024. He has been a Ph.D. candidate in the School of Transportation Science and Engineering, Harbin Institute of Technology, Harbin, China since 2024. His research interests include ground penetrating radar (GPR) detection and processing of its data.

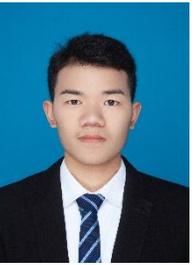
**Yuhui Zhang** received the B.S. degree in Road, Bridge and Cross River Engineering from Zhengzhou University, Henan, China, in 2023. He is currently working toward the M.S. degree in Traffic and Transportation Engineering with the School of Transportation Science and Engineering, Harbin Institute of Technology, Harbin, China. His research interests include non-destructive testing, ground penetrating radar (GPR) and data fusion.

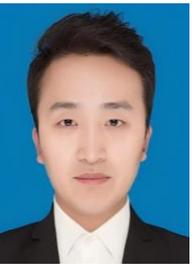
**Zepeng Fan** received the Ph.D degree from RWTH Aachen University in 2021. He is currently an associate professor of School of Transportation Science and Engineering, Harbin Institute of Technology, Harbin, China. His research interests include multiscale modelling and characterization of bitumen-aggregate interfacial behavior, and development of sustainable and low carbon pavement materials and technologies. He has published more than 50 SCI papers and served as a reviewer for more than 20 SCI journals. He has undertaken over 8 national and provincial projects.

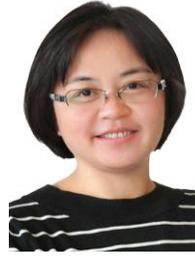
**Yiqiu Tan** received the Ph.D. degree in transportation. She is currently a Full Professor with the School of Transportation Science and Engineering, Harbin Institute of Technology. She is also the Executive Deputy Director of the Key Laboratory of the Transportation Industry. She has long been engaged in research on the basic theory and applied technology of pavement structure and materials. She has published more than 200 academic articles, four monographs, and 47 authorized invention patents. Her research results have produced enormous economic, social, and environmental benefits.

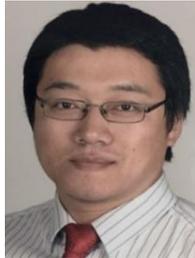
**Dawei Wang** (Senior Member, IEEE) received the B.S. in civil engineering in Tsinghua University in 2003, M.S. and Ph.D. in roadway engineering in RWTH Aachen University in 2007 and 2011, respectively. In 2017, he was granted Habilitation based on the research and effort he has contributed to the highway engineering in Germany. His research interests and expertise focus primarily on asphalt pavement skid resistance, multi-scale characterization of the asphalt pavement mechanical behavior and functional pavement theory and technology. So far, he has directed more than 24 scientific research projects. Over the last years, he has published nearly 200 academic publications, including more than 140 SCI indexed papers. He also serves on the editorial boards of many international academic journals.

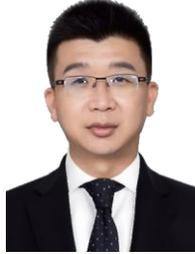
**Binglei Xie** received the B.S. degree in applied mathematics and the Ph.D. degree in management science and engineering from Southwest Jiaotong University, Chengdu, China, in 1996 and 2003, respectively. He is currently a professor of transportation planning and management with the School of Architecture, Harbin Institute of Technology (Shenzhen), Shenzhen, China. His primary research interests are travel behavior analysis, emergency transportation management, and integrated land use/transport planning. His published articles appear in Transportation Research Part E and Advanced Engineering Informatics, among others.